\documentclass[nohyperref]{article}

\usepackage{microtype}
\usepackage{graphicx}
\usepackage{subfigure}
\usepackage{booktabs} % for professional tables

\usepackage{hyperref}

 \usepackage[accepted]{icml2023}

\usepackage{amsmath}
\usepackage{amssymb}
\usepackage{mathtools}
\usepackage{amsthm}

\usepackage[capitalize,noabbrev]{cleveref}

\theoremstyle{plain}

\theoremstyle{definition}

\theoremstyle{remark}

\usepackage[justification=centering]{caption}% or e.g. [format=hang]
\usepackage[verbose]{placeins}

\usepackage{color}
\definecolor{dkgreen}{rgb}{0,0.65,0}
\definecolor{gray}{rgb}{0.5,0.5,0.5}
\definecolor{mauve}{rgb}{0.8,0,0.5}
\usepackage{listings}
\newcommand{\x}{\bold{x}}
\newcommand{\z}{\bold{z}}
\newcommand{\boldc}{\bold{c}}
\newcommand{\wmu}{w_\mu}
\newcommand{\wsigma}{w_\sigma}
\newcommand{\bmu}{\boldsymbol{\mu}}
\newcommand{\bsigma}{\boldsymbol{\sigma}}

\newcommand{\pxz}{p(\x \vert \z)}
\newcommand{\qsingle}{q(\z \vert \x)}
\newcommand{\posterior}{q(\z_i \vert \z_{<i}, \x)}
\newcommand{\prior}{p(\z_i \vert \z_{<i})}
\newcommand{\condprior}{p(\z_i \vert \z_{<i}, \boldc)}
\newcommand{\brackets}[1]{\left[#1\right]}
\newcommand{\pars}[1]{\left(#1\right)}
\newcommand{\Expect}[2]{\mathbb{E}_{#1}\brackets{#2}}
\newcommand{\KLdiv}[2]{\text{KL}({#1} \Vert {#2})}
\newcommand{\dz}{\text{d}\z}

\usepackage[textsize=tiny]{todonotes}

\begin{document}

\twocolumn[
\icmltitle{High Fidelity Image Synthesis With Deep VAEs In Latent Space}



\icmlsetsymbol{equal}{*}

\begin{icmlauthorlist}
\icmlauthor{Troy Luhman}{equal}
\icmlauthor{Eric Luhman}{equal}

\end{icmlauthorlist}

% \icmlkeywords{Machine Learning, ICML, Computer Vision, Generative Modeling, VAE}

\icmlcorrespondingauthor{}{troyluhman@gmail.com}
\icmlcorrespondingauthor{}{ericluhman2@gmail.com}

\vskip 0.3in
]



\printAffiliationsAndNotice{\icmlEqualContribution} % otherwise use the standard text.

\begin{abstract}
We present fast, realistic image generation on high-resolution, multimodal datasets using hierarchical variational autoencoders (VAEs) trained on a deterministic autoencoder's latent space. In this two-stage setup, the autoencoder compresses the image into its semantic features, which are then modeled with a deep VAE. With this method, the VAE avoids modeling the fine-grained details that constitute the majority of the image's code length, allowing it to focus on learning its structural components. We demonstrate the effectiveness of our two-stage approach, achieving a FID of 9.34 on the ImageNet-256 dataset which is comparable to BigGAN. We make our implementation available online\footnote{
Torch code: \url{https://github.com/ericl122333/latent-vae}, Jax code: \url{https://github.com/ericl122333/latent-vae-jax}}.
\end{abstract}
%rewrite this

\section{Introduction}
\label{intro}

There has recently been a surge of interest in deep generative modeling for image generation, which has made great progress in the last few years. In particular, advances in diffusion and autoregressive models have enabled them to create photo-realistic images for even the most complex object compositions \citep{dalle2, ldm, parti}. Generative models like these will likely have significant impact on many creative applications as they become continually adopted and improved.

Despite their impressive capabilities, both these model classes have a serious drawback which is their slow sampling speed. Producing an image generally requires evaluating a large neural network 50-1000 times since each transition in their generative process reuses the function. This inefficiency is an important consideration in practical settings, and can pose a challenge to widespread deployment of the model.

Variational Autoencoders (VAEs) are a class of likelihood-based models that use learned latent variables to generate data \citep{originalvae, rezende}. Hierarchical VAEs separate latents into multiple conditionally dependent groups for greater expressitivity \citep{laddervae, iafvae}. Notable examples of hierarchical VAEs for images include NVAE \citep{nvae} and VDVAE \citep{vdvae}. Unlike diffusion and autoregressive models which use a fixed analytic posterior, deep VAE posteriors are trainable and parameterized by neural networks. Their learned ordering of latent variables results in an efficient generation process that only uses one function evaluation. Beyond their fast sampling speed, hierarchical VAEs are also desirable due to their easy inference, stable training, and interpretable latent representations. 

However, to date hierarchical VAEs have not suceeded in generating convincing images on large, multimodal datasets like ImageNet \citep{imagenet}. This is especially surprising given that their hierarchical generation process seems naturally suited for image generation. Indeed, autoregressive models have proven far more successful despite a counter-intuitive inductive bias that simply generates images in raster-scan order. Thus, we start by observing ingredients that make autoregressive image models so successful and ask whether similar techniques can be applied to VAEs.

\begin{figure*}
  \centering \includegraphics[scale=0.29]{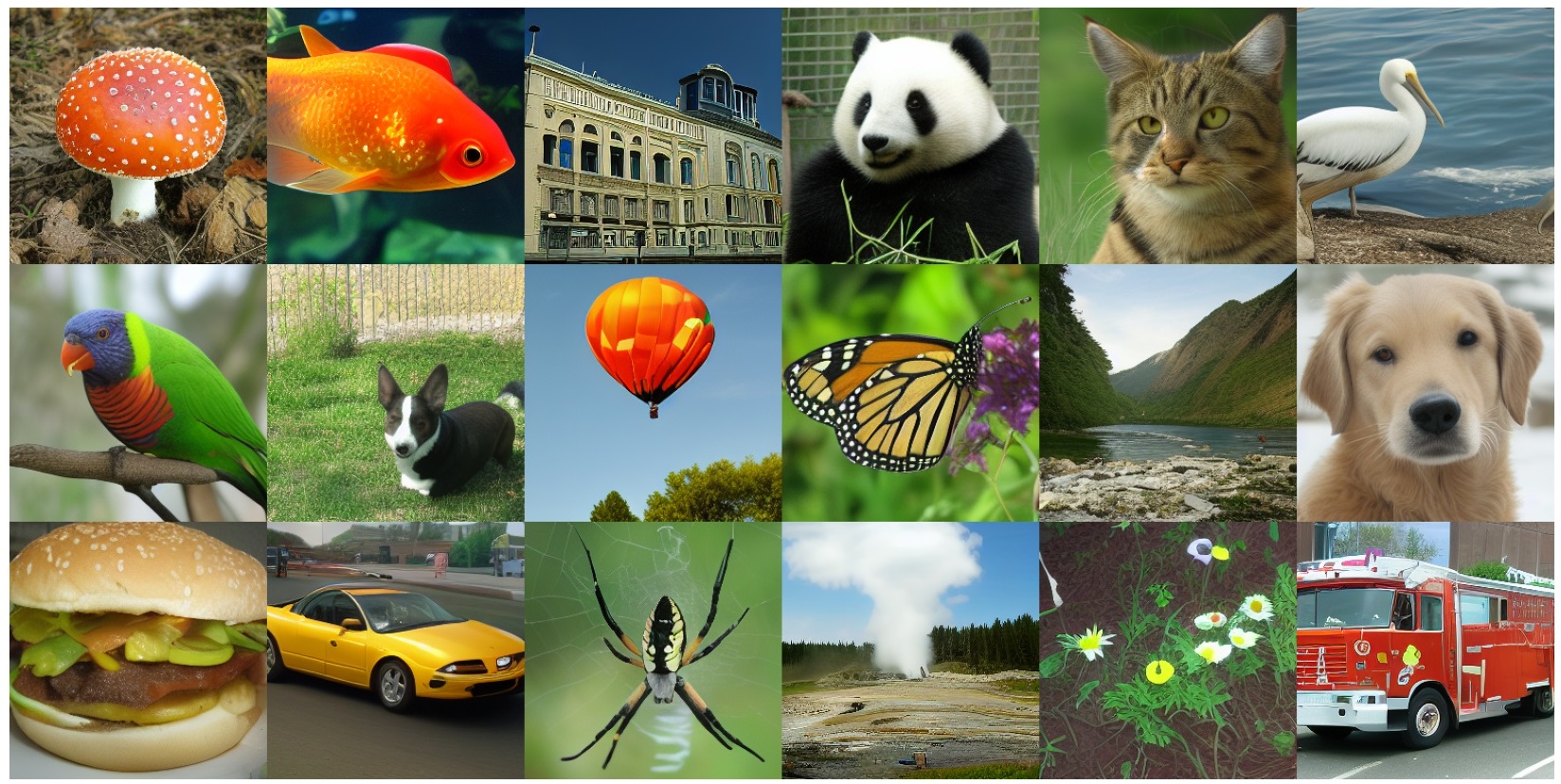}
  \vspace{-0.2cm}
  \caption{Selected samples from our guided ImageNet-256 model (FID=9.34).}
\end{figure*} 

A crucial component to the success of autoregressive models is training on a sequence of compressed image tokens instead of direct pixel values \citep{vqvae, vqgan}. Doing so enables them to focus on learning the relationships between image semantics while ignoring imperceptible image details, which is especially important because imperceptible details make up a great majority of an image's code length \citep{pope2021intrinsic, ddpm, ldm}. We hypothesize that, like pixel-space autoregressive models, existing pixel-space hierarchical VAEs focus primarily on learning fine-grained features, weakening their ability to learn the underlying composition of image concepts. 

Motivated by this theory, we propose to improve deep VAEs by applying them to the latent space of a deterministic autoencoder (DAE). Our pipeline consists of two parts trained in succession: a DAE that reconstruct images from low-dimensional latents, and a VAE that forms a generative model of these latents. Compared to acting directly on pixel-space, training the VAE on low-dimensional latents has two major advantages. Firstly, the code length of the compressed latent is much smaller than its RGB representation, but it preserves nearly all the image’s perceptual information. Having a smaller total code length is helpful because it makes global features, which make up only a few bits, more strongly emphasized. Imperceptible details which would otherwise be modeled by the VAE are now discarded, allowing the VAE to focus solely on image structure. Secondly, the latent variable has greatly reduced dimensionality, which reduces computational costs and allows for larger models to be trained with the same resources.

We evaluate our method on several image benchmarks, and obtain strong performance that greatly exceeds previous hierarchical VAE capabilities. When combining the proposed two-stage method with classifier-free guidance on ImageNet-256, we obtain a FID score \citep{fid} score of 9.34 that is competitive landmark diffusion and GAN baselines \citep{biggan, adm}. Additionally, we experiment with VAE's latent representation to showcase their good interpretability and flexibility.

\section{Related Work}
\label{section:2}

\textbf{Hierarchical VAEs}. 
Our work builds off previous literature in VAEs \citep{originalvae, rezende}, and their hierarchical extensions \citep{laddervae, iafvae, biva, nvae}. In these hierarchical VAEs, early latent groups start at low dimensionality and increase in dimensionality over the course of generation. For image data, this progression from low to high resolution naturally leads to a coarse to fine generation process. These hierarchical VAEs benefit from having many latent groups, where \cite{vdvae} showed increasing statistical depth substantially improves likelihood. A downside of higher depth is the increased compute and memory costs, which can impede scaling to high-resolution data. Our method addresses this by using a shallow DAE for high-frequency relationships, while learning complex semantic relationships with a deep VAE.

Compared to other likelihood-based models, VAEs are distinguished by their powerful posteriors parameterized by neural networks. This enables them to learn an efficient ordering of latent variables and results in fast sampling speed. In contrast, diffusion and autoregressive models train a powerful prior to match a simple, hand-crafted posterior. The neural network parameterizing the prior shares parameters across latent layers, and its repeated evaluation at sampling time leads to slow generation. Normalizing flows \citep{rezende2015variational, realnvp} are a likelihood-based model with fast sampling speed, but the need for explicit invertibility weakens their modeling capabilities in practice.

\textbf{Two-stage Image Generation}. 
Instead of directly generating high-resolution images, many works have found dividing the generation process into two stages is beneficial for both efficiency and performance \citep{vqvae2, vqgan, cdm, ldm}. One leading method is to train an autoregressive model on discrete image tokens produced by an encoder neural network \citep{vqvae, vqgan, vitvqgan}; this sequence-like tokenization is a helpful inductive bias for the transformer architecture these models use. VQ-VAEs train a discrete VAE for the first-stage encoder, and VQ-GANs achieve sharper images by adding auxiliary perceptual and adversarial losses to the reconstruction objective \citep{vqgan}. \cite{ldm} also trains an encoder with the same loss objectives, but it outputs continuous values and trains a diffusion model over these latents. Our work adapts the pretrained autoencoders from \cite{ldm} since our VAEs act in continuous space. 

An alternative two-stage approach uses an upsampling stack which first creates low-resolution RGB images before upsampling them to a higher resolution \citep{saharia2022image, cdm}. This method works very well for diffusion models \citep{adm, dalle2} as diffusion models have good super-resolution capabilities \citep{li2022srdiff, palette}. In our hierarchical VAE experiments, we found that upsampling stack is outperformed by a latent-space approach (Section \ref{section:4.3}).

Our work is perhaps most similar to \cite{dai2019diagnosing}, who propose a VAE on the latent space of another VAE learned directly on pixels. This work is similarly motivated to ours, accounting for the fact that the ambient dimension is much greater than the manifold dimension. However, samples from their model are blurry since their first stage is a regular VAE trained with a pixel-space distortion term. In contrast, we learn the underlying manifold using a DAE augmented with perceptual and aversarial losses, and achieve much higher compression rates with little to no blurriness. Furthermore, our second stage is based off hierarchical VAEs instead of single-group ones to reap the benefits of high stochastic depth observed in \cite{vdvae}.

\textbf{Perceptual and Adversarial Losses in VAEs}. 
Previous works \citep{vaegan, perceptualvae} have found that the pixel-wise reconstruction objective in VAEs encourages them to produce blurry, unrealistic samples. As a result, they modify the reconstruction objective to include a perceptual or adversarial loss, which encourages the reconstructions to be visually similar to real images in the feature space of a deep CNN. While both our method and theirs include perceptual or adversarial losses in the reconstruction objective, our adversarial component is trained separately from the MLE-based VAE component, whereas these works combine the two objectives in a joint training scheme. 

\section{Method}
\label{section:3}

\subsection{Perceptual Compression Stage}
\label{section:3.1}

In a two-stage image generation approach, the first step often involves training an autoencoder that compresses high-resolution images onto a lower-dimensional latent space. Specifically, a neural network encoder takes in an image $\x \in \mathbb{R}^{H \times W \times 3}$ and outputs a latent code $\textbf{h} \in \mathbb{R}^{H/f \times W/f \times C}$ for some downsampling factor $f$ and output channel size $C$. From this, a decoder tries to reconstruct the original image $\x$ as best as possible. The autoencoder's objective combines an L1 loss with perceptual and patch-based adversarial losses \citep{vqgan, ldm}. Perceptual and adversarial losses are crucial to attaining sharp reconstructions, since the decoder would otherwise produce a blurry average of prediction to minimize pixel-space distortion. Meanwhile, instabilities caused by adversarial training are tempered by the L1 and perceptual losses.

Our motivation for the autoencoder stems from the manifold hypothesis \citep{deeplearning}, which states that real-world data modalities lie on a low dimensional manifold in a high dimensional space. This implies that the autoencoder can learn representations with much lower dimensionlity but contain all factors of variation in the data. Such representations might be beneficial to a generative model for several reasons. One reason could be the excessive amounts of trivial relationships present in high-dimensional data. These perceptually meaningless relationships make up many of the dependencies in the ambient space, so removing them allows focusing on the most important concepts. Additionally, data distributions in high dimensional spaces tend to  be extremely sharp, with data modes separated by wide boundaries with no density at all. Reducing this sharpness could be especially beneficial for VAEs which must fit Guassian posteriors to Gaussian priors. %should we add a citation for the "estimating manifold dimension" by pope et al 2021 somewhere here? to defend some of the claims earlier in the page

\subsection{Hierarchical VAEs}
\label{section:3.2}

Hierarchical VAEs are latent variable models that train a generator $p(\x \vert \z)$ to reconstruct data $\x$ from a sequence of latent variables $\z \coloneqq \{\z_1, \ldots, \z_N \}$ sampled from a learned posterior. The posterior encodes information about the data necessary for reconstruction, but is also incentivized to be close to a prior distribution. The posterior and prior are generally factorized normal distributions of the form $\qsingle \coloneqq \prod_{i=1}^{N} \posterior$ and $p(\z) \coloneqq \prod_{i=1}^{N} \prior$ respectively. The generator, posterior, and prior are trained jointly to maximize a lower bound on the data log-likelihood:

\begin{multline}
\label{equation:1}
\log p(\x) \geq \Expect{\qsingle}{\log \pxz} - \KLdiv{q(\z_1 \vert \x)}{p(\z_1)} \\ -\sum_{i=2}^{N}\Expect{q(\z_{<i} \vert \x)}{\KLdiv{\posterior}{\prior}}
\end{multline}

The objective in Equation \ref{equation:1} is analogous to minimizing a lossless codelength of the data, with the reconstruction loss describing the distortion measured in nats, and KL terms representing the rate \citep{vlae}. However, if most bits in a natural image encode barely perceptible details, then this lossless compression task is dominated by the perceptual compression component. In other words, a VAE trying to minimize its loss would do so by mastering high-level features while ignoring the image's semantics, since the former make up the majority of the image's codelength. 

This problem disappears when training in the latent space since high-frequency details have already been removed by the autoencoder. Thus to minimize its loss, a latent-space VAE must learn the complex dependencies between image concepts. Fortunately, it can dedicate all its capacity to this single task without spending any on learning details. This is especially important for deep VAEs, since it redistributes not only parameters but also latent groups.

\subsection{Classifier-Free Guidance}
\label{section:3.3}

In addition to learning on lower dimensional representations, large-scale diffusion and autoregressive models employ classifier-free guidance \citep{cfg} to boost image fidelity \citep{glide, imagen, makeascene, parti}. This technique is designed to trade off diversity for sample quality, since poor likelihood-based models tend to cover regions not present in the data distribution. Guidance pushes samples towards regions that more closely resemble a desired label by comparing conditional and unconditional likelihood functions. 

\cite{visvae} extends this technique to hierarchical VAEs where means and log-variances are extrapolated towards class-conditional values and away from unconditional ones. This VAE guidance strategy considers diagonal Gaussian priors $\prior$ conditioned only on previous latents, and priors $\condprior$ conditioned on both previous latents and an auxiliary label $\boldc$. The guided latent variables are drawn from $p_{\text{guided}, \wmu, \wsigma}(\z_i \vert \z_{<i}, \boldc)$ defined as: 
\begin{multline}\label{equation:2}
p_{\text{guided}, \wmu, \wsigma}(\z_i \vert \z_{<i}, \boldc) \coloneqq \\
 \mathcal{N} \pars{\z_i; \bmu_c + \wmu(\bmu_c - \bmu_u), \text{diag} \pars{\bsigma_c^2 \pars{\frac{\bsigma_c}{\bsigma_u}}^{2\wsigma}} }
\end{multline}
for conditional means and covariances $\bmu_c, \text{diag}\pars{\bsigma_c^2}$ and unconditional ones $\bmu_u, \text{diag}\pars{\bsigma_u^2}$. The strength of the mean and variance guidance are controlled by $\wmu$ and $\wsigma$ respectively, which may be chosen independently. 

Intuitively, mean guidance accentuates components that are present in the class-conditional value but not in the unconditional one. This helps give generated images a distinct visual structure that is characteristic of other images in the class. Variance guidance acts similarly to low-temperature sampling, adaptively adjusting the temperature based on how the ratio of class-conditional variance compares to the unconditional one. 

To implement guidance in VAEs, we train a model where all priors are conditioned on a label that is replaced with a dummy label 10\% of the time. At sampling time, we store running hidden states for both the conditional and unconditional generation paths. Each path outputs its own prior, and we combine the two into a single guided distribution according to Equation \ref{equation:2}. A single sample $\z \sim p_{\text{guided}}$ is then projected back into both hidden states (see Appendix \ref{appendix:C} for a PyTorch implementation).

\section{Experiments}
\label{section:4}

In this section, we ablate the effect of training in the latent space for various image benchmarks, and evaluate our method's image generation capabilities and its efficiency. Full details of our setup is listed in Appendix \ref{appendix:B}. For each dataset considered in this section, we use an off-the-shelf autoencoder from \cite{ldm} as our first stage model. 

\subsection{Ablation Studies}
\label{section:4.1}

We first test whether training in the latent space improves over a pixel space VAE on the LSUN Church-256 dataset. We also compare to an upsampling stack, where a deep VAE is trained to generate low-resolution images, which are then upsampled with a second VAE. Both latent and upsampling methods produce $32^2$ outputs in the first stage, and have 119 million parameters. Our VAE upsampler has 63 million, while our DAE upsampler has slightly more at 80 million. All models are trained for 250 thousand steps and roughly the same wall clock time ($\sim 350$ TPUv2-8 hrs).

\setlength{\tabcolsep}{4.5pt}
\begin{table}[h!]
  \begin{center}
    \caption{Comparison between methods on LSUN Church-256.} 
    \label{tab:table1}
    \begin{tabular}{lcccc} 
      Method&FID $\downarrow$ &Prec $\uparrow$ &Rec $\uparrow$\\
      \hline \\
      Latent-space VAE &7.89 &0.69 &0.35\\
      Upsampling Stack &33.53 &0.58 &0.10\\
      Pixel-space VAE &44.36 &0.48 &0.13\\
    \end{tabular}
  \end{center}
\end{table}

Table \ref{tab:table1} shows the ablation results. The latent-space VAE obtains much better FID than the upsampling stack and pixel-space VAE, performing comparably to a diffusion model from \cite{ddpm}. The good FID and precision verify the latent VAE's more realistic samples, but its higher recall indicates better mode coverage of the data manifold. The pixel-space VAE covers modes in the ambient space better because of its data log-likelihood objective, but its samples' lack of realism make them fall outside the data manifold. The compressed DAE representation more closely resembles this manifold, allowing useful learning signal to flow into the generative model.

Next, we ablate the effect of the downsampling factor on performance. Three models with $f=4$, $f=8$ and $f=16$ downsampling are trained on the LSUN-Bedroom dataset. All models have $\sim 100$ million parameters and trained for 250 thousand steps. 

\setlength{\tabcolsep}{4.5pt}
\begin{table}[h!]
  \begin{center}
    \caption{Comparison between downsampling factors on LSUN Bedroom-256} 
    \label{tab:table2}
    \begin{tabular}{lcccc} 
      Downsampling Factor &FID $\downarrow$ &Prec $\uparrow$ &Rec $\uparrow$\\
      \hline \\
      $ f = 4$ &9.35 &0.56 &0.36\\
      $ f = 8$ &11.16 &0.47 &0.36\\
      $ f = 16$ &17.46 &0.38 &0.28\\
    \end{tabular}
  \end{center}
\end{table}

As shown in Table \ref{tab:table2}, a compression rate of $f = 4$ performs only slightly better than $f = 8$, but $f = 16$ performs notably worse. This indicates that the benefit of dimensionality reduction has diminishing returns beyond a certain point, and excessive compression rates such as $f = 16$ can hurt overall performance. Our observations are consistent with \cite{ldm}, who find that compression rates of $f = 4$ and $f = 8$ work best with diffusion models.

\subsection{Quantitative results}
\label{section:4.2}

Table \ref{tab:table3} evaluates our method on the class-conditional ImageNet $256^2$ dataset using the Frechet Inception Distance (FID) and Inception Score (IS) metrics \citep{fid, inceptionscore}, as well as the Precision and Recall metrics to assess fidelity and diversity respectively [precrecall]. We train a 386M parameter model for 475k iterations on a batch size of 1024, with an $f=8$ autoencoder from \cite{ldm} as our first stage.

We obtain a FID of 9.34 when performing guided sampling with $\wmu = 1.5$ and $\wsigma = 3$\footnote{Mean guidance not used for $1\times1$ latents to reduce saturation.}. Surprisingly, this result is only slightly worse than LDM-8, which was with the same methodology and downsampling factor but uses a diffusion backbone. This provides evidence that the gap between diffusion models and VAEs greatly shrinks when moving to a compressed latent space. 

When not applying guidance, our VAE's sample quality degrades significantly as measured by FID and Recall. Our strong performance in the guided setting indicates useful features have been learned, but the noisiness of unguided sampling prevents this from being translated into realistic images. Guidance directs the model to regions where it can effectively utilize these learned features. Nevertheless, our unguided samples exhibit high recall, implying diverse mode coverage. Even if these samples are not as impressive visually, reliably covering modes is a desirable feature and an advantage of VAEs over GANs. 

\setlength{\tabcolsep}{3.5pt}
\begin{table}[h!]
  \begin{center}
    \caption{Performance on ImageNet $256^2$. $\dagger$ indicates methods using classifier-free guidance. $\ddagger$ indicates two-stage approaches. All numbers taken from \cite{adm} except for LDM and StyleGAN-XL, which we take from their corresponding papers.} 
 
    \label{tab:table3}
    \begin{tabular}{lcccc} 
      \\ 
      Method&FID$\downarrow$ &Prec$\uparrow$ &Rec$\uparrow$\\
      \hline \\
      Latent VAE$^{\dagger \ddagger}$ (ours, guided) &9.34 &0.85 &0.32\\
      Latent VAE$^{\ddagger}$ (ours, unguided) &32.7 &0.56 &0.59\\
      VQVAE-2$^{\ddagger}$ \citep{vqvae2} &31.11 &0.36 &0.57\\
      LDM-8 $^{\dagger \ddagger}$ \citep{ldm} &8.11 &0.83 &0.36\\
      BigGAN-deep \citep{biggan} &6.95 &\textbf{0.87} &0.28\\
      LDM-4$^{\dagger \ddagger}$ \citep{ldm} &3.60 &\textbf{0.87} &0.48\\
      ADM \citep{adm} &10.94 &0.69 &\textbf{0.63}\\
      ADM-G$^{\dagger}$ \citep{adm} &4.59 &0.82 &0.52\\
      StyleGAN-XL \citep{styleganxl} &\textbf{2.30} &0.78 &0.53
      
    \end{tabular}
  \end{center}
\end{table}

\subsection{Qualitative results}
\label{section:4.3}

\begin{figure}[hbt!]
\begin{center}
\includegraphics[scale=0.16]{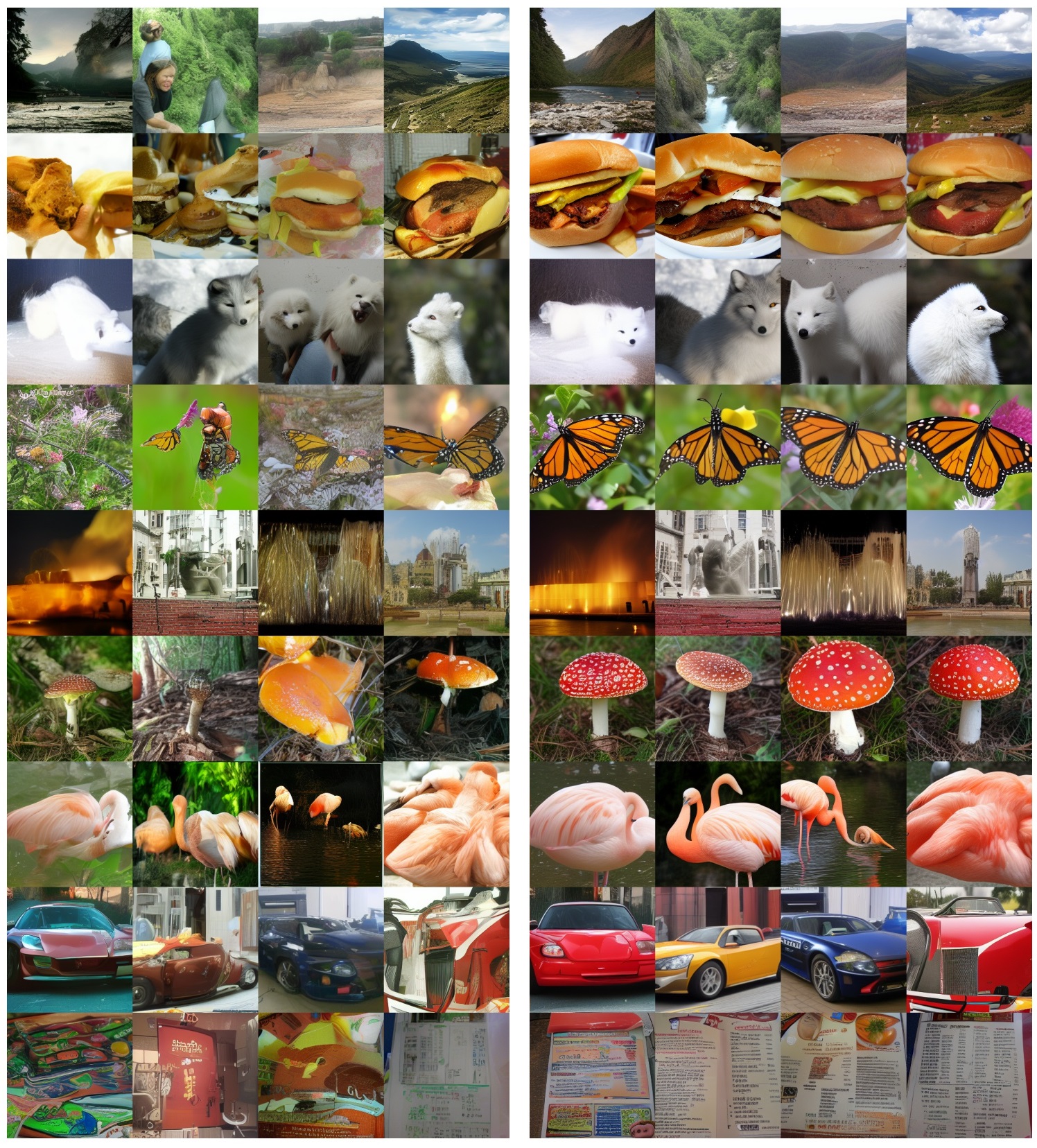}
\caption{Class-conditional samples from our ImageNet model without guidance (left, FID 32.7), and with guidance (right, FID 9.34). Guidance significantly improves the realism of the generated samples.}\label{fig:2}
\end{center}
\end{figure}

Figure \ref{fig:2} shows random samples from our ImageNet model. The guided samples demonstrate VAEs' ability to create realistic pictures with both coherent global structures and realistic textures. Although the unguided samples are less realistic, they have greater variety in their composition and backgrounds, demonstrating VAEs ability to trade sample diversity for perceptual quality. The effect of guidance also differs from image to image: the landscape in the last column of the first row changes little when applying guidance, while the photos of butterflies are changed drastically.

\begin{figure}[hbt!]
\begin{center}
\includegraphics[scale=0.145]{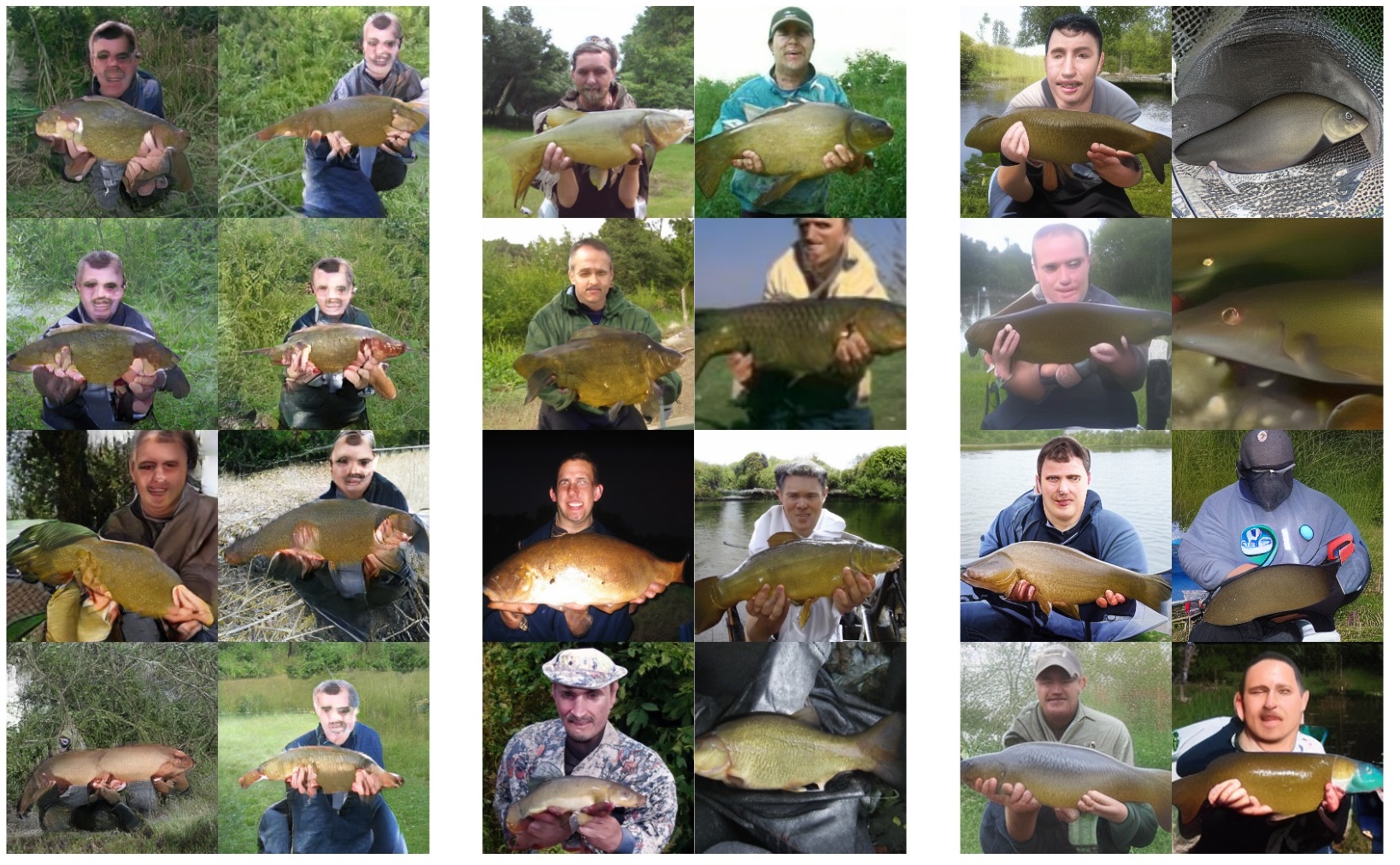}
\caption{Examples for the tinca class from BigGAN-deep (left), ADM (center), and ours (right). Like diffusion models, VAEs show significant improvement over GANs at generating faces.}\label{fig:3}
\end{center}
\end{figure}

In Figure \ref{fig:3}, we compare samples from our model, BigGAN, and ADM for the tinca class. Our model is capable of generating diverse settings, such as tincas not held by a person. Furthermore, our model renders human faces more realistically than BigGAN. The diversity of model samples and realistic generation of difficult concepts show how VAEs emulate the most desirable aspects of diffusion models. Finally, we visualize nearest neighbors in Appendix \ref{appendix:D} to demonstrate that our model does not simply memorize the training set.

\subsection{Efficiency}
\label{section:4.4}

Similar to GANs and unlike diffusion models, VAEs perform their entire generative process in a single forward pass. As such, we would expect VAEs to vastly outperform diffusion models in sampling speed. Table \ref{tab:table4} supports this claim empirically by comparing our VAE's efficiency to common diffusion and GAN methods. Our method is similarly fast as StyleGAN-XL \citep{styleganxl}, while boasting a $20\times$ speedup over a 50-step LDM and a $300\times$ speedup over ADM \citep{ldm, adm}. 

\setlength{\tabcolsep}{3.5pt}
\begin{table}[h!]
  \begin{center}
    \caption{Sampling time (s) on ImageNet-256, measured for 1000 images on a Tesla T4 GPU. Guidance is used when applicable.} 
    \label{tab:table4}
    \begin{tabular}{lcc} 
      \\ 
      Method & Sampling Time & Speedup Factor\\
      \hline \\
      Latent VAE (ours) & 213.8 & 1$\times$ \\
      StyleGAN-XL & 272.9 & 0.781$\times$ \\
      LDM-4 (50 steps) & 4560 & 0.047$\times$ \\
      ADM (250 steps) & 72875 & 0.003$\times$ \\
      
    \end{tabular}
  \end{center}
\end{table}

Although VAEs have an inherently fast sampling process, existing VAEs are notoriously expensive to train \citep{harvey2021conditional}. This inefficiency stems from the need to back-propagate through a deep hierarchy of latent variables. As an example, a VD-VAE on FFHQ-256 uses 62 latent layers and roughly 1000 convolutional layers, and takes over 300 V-100 days of compute to train to completion \citep{vdvae}. To present a more complete picture of our method's efficiency, we report the training speed of our method. 

Our base ImageNet generator was trained for 26 days on TPUv3-8 hardware, which equates to only 104 days on a single V-100. Even adding $\sim 70$ V-100 days for training the autoencoder, this training time is significantly better than VD-VAE's, despite our generator having three times more parameters. We achieve this fast training speed by only applying deep VAE layers only to low-resolution components, using a wide but shallow autoencoder at high-resolutions.

\section{Image Manipulations}
\label{section:5}

A distinguishing feature of VAEs is their powerful inference network where data points are projected onto semantically meaningful latent space. This allows for both good interpretability as well as flexible manipulation. Section \ref{section:5.1} examines learned representations using visualizing two latent dimensions in a grid, and Section \ref{section:5.2} creates interpolation paths between two dataset images. Section \ref{section:5.3} showcases how latents can be manipulated to perform image outpainting without any task-specific training or auxiliary networks. For simplicity, this section considers only our unconditional models, which were trained with a standard normal prior.

\subsection{Visualizing a Manifold}
\label{section:5.1}

One desirable aspect of a latent variable model is good interpretability. This lets researchers and practitioners understand what representations have been learned and how they affect the model's decisions. The seminal work of \cite{originalvae} used a VAE with a two-dimensional latent code to visualize the entire learned manifold of a small-scale dataset. While our models and datasets are far more complex and high-dimensional, we are interested in understanding the learned latent representations.

\begin{figure}[hbt!]
\begin{center}
\includegraphics[scale=0.14]{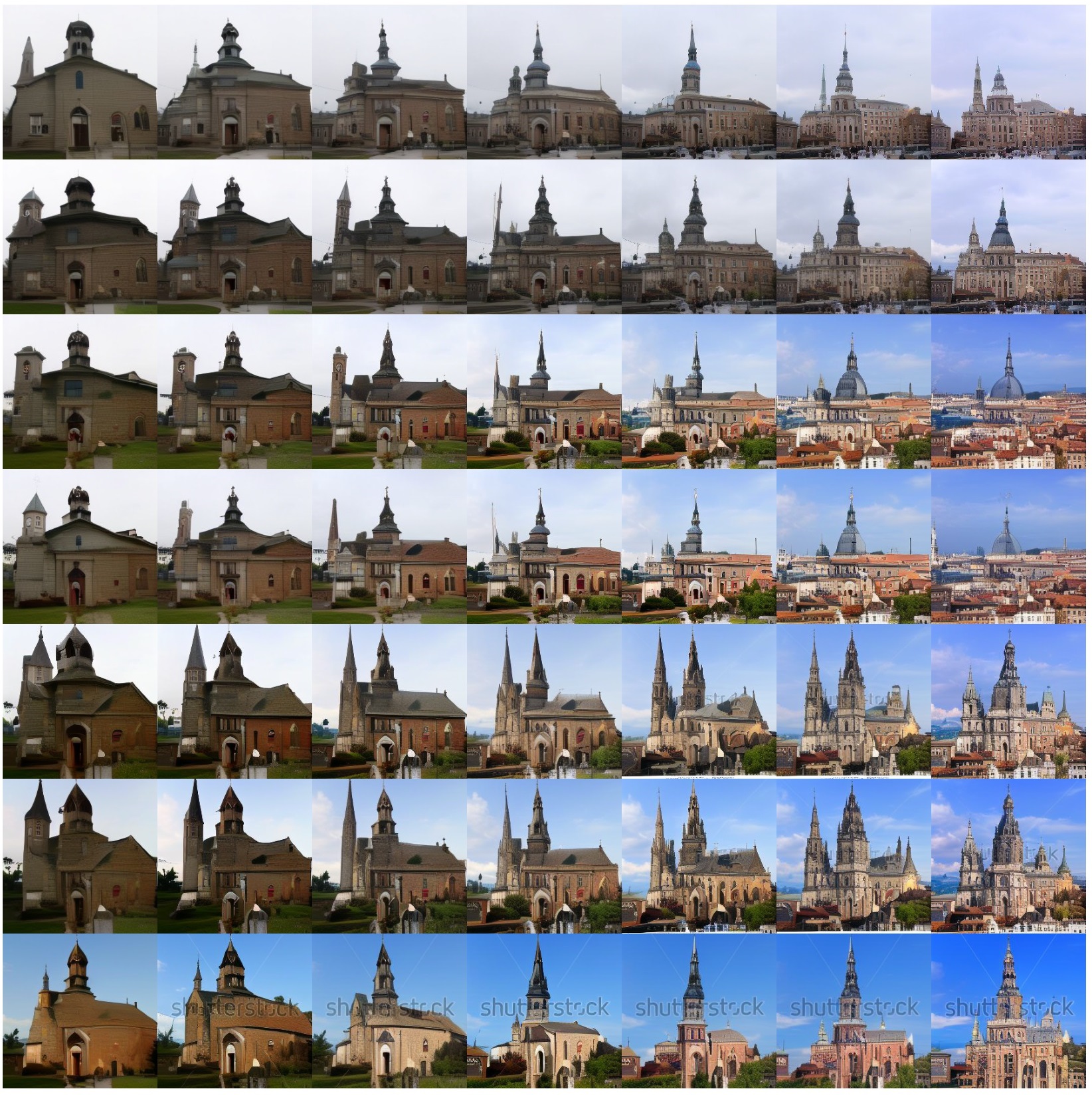}
\caption{Traversing two dimensions of the learned latent space. We vary the noise seed for the most informative dimension of the first two layers, with linearly spaced intervals from -2.25 to 2.25.}
\label{fig:4}
\end{center}
\end{figure}

Our deep VAEs have thousands of latent dimensions starting from $1 \times 1$ resolution all the way to $32 \times 32$. Despite this high dimensionality, most global features are encoded in $1 \times 1$ and $4 \times 4$ latents \citep{vdvae}. To retrieve a two-dimensional representation, we consider the first two latent groups at the $1 \times 1$ resolution. Each of these groups has 16 channels each, but similar to \cite{efficientvdvae}, we find that most the information is stored in only one of these 16 channels. We create a grid by sweeping over the most active channel in the first two latent groups, sampling all other dimensions with a fixed noise seed.

Figure \ref{fig:4} shows how these two dimensions affect samples of our LSUN Church model. The churches become progressively more intricate when moving left to right, while moving top to down changes the image's vibrance and color palette. The structure of the church stays relatively constant as we move across these two dimensions, indicating this information is encoded in later latent variables. 

\subsection{Interpolation}
\label{section:5.2}

Another way of examining learned representations is by interpolating between different samples. We are interested in generating between real images from the dataset instead of synthetic model samples. For GANs, this is a more complicated problem which requires either an auxiliary encoding network or multiple optimization steps. However, in VAEs the inference network is trained directly in the objective and is directly applicable to downstream tasks.

A simple way of interpolating between images is to simply interpolate between the latent codes that created them. In other words, one could sample $\z^{(1)} \sim q(\z \vert \x^{(1)})$ and $\z^{(2)} \sim q(\z \vert \x^{(2)})$ to produce an interpolated latent $\z_{\text{interp}} \coloneqq \lambda \z^{(1)} + (1 - \lambda) \z^{(2)}$. The interpolated latent is converted to an image using the decoder $p(\x \vert \z_{\text{interp}})$. 

\begin{figure}[hbt!]
\begin{center}
\includegraphics[scale=0.13]{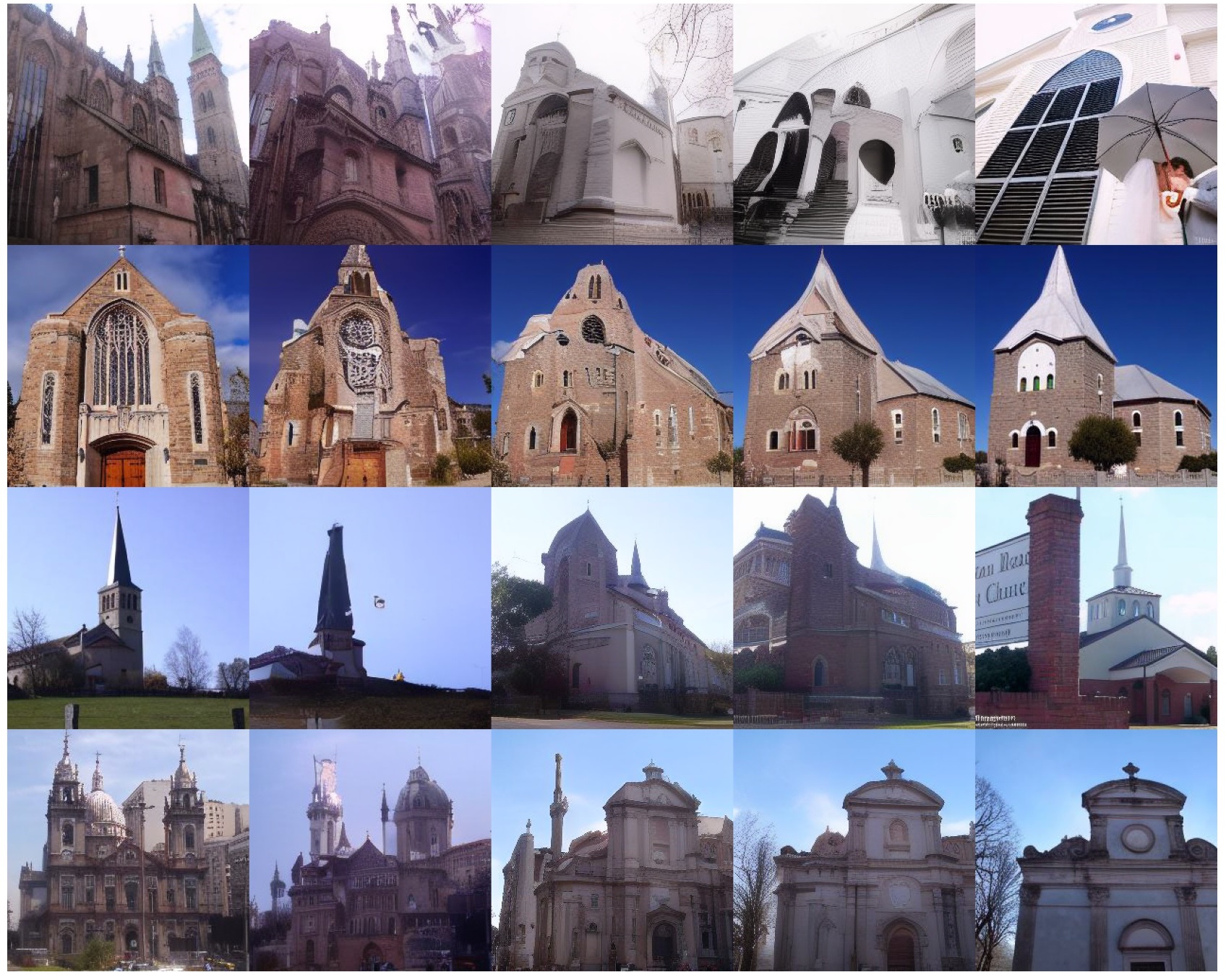}
\caption{Interpolation results. Leftmost and rightmost columns are samples from the dataset, while the middle three columns are generated interpolations.}
\label{fig:5}
\end{center}
\end{figure}

Figure \ref{fig:5} shows interpolations between samples from the LSUN Church-256 dataset. Despite its simplicity, this interpolation method works for models with a standard normal prior. This makes sense because each posterior is trained to fall within the same prior, so interpolating between points sampled from these posteriors should produce another point within the prior. In models with hierarchical priors, this direct interpolation fails because the interpolated latent falls outside the priors learned during training. It may be possible to reparameterize the hierarchical prior into a standard normal one, as done in Appendix E of \cite{sbmlatent}, but we leave that to future work.

\subsection{Outpainting}
\label{section:5.3}

We now explore how to manipulate latent variables for an image outpainting task. Outpainting involves generating a larger image while its context is provided by a reference image. Although our models are not explicitly conditioned with the context provided, we can store this information in latent variables that are used alongside newly sampled latents. Latents corresponding to outpainted regions are sampled from a standard normal prior, while regions corresponding to the original image use latents from the posterior.

To outpaint a $256^2$ image into $256 \times 512$ image, we perform the following steps. First, we encode the $256^2$ image which outputs a latent code with $1 \times 1$, $4 \times 4$, $8 \times 8$, $16 \times 16$, and $32 \times 32$ resolutions. We then double the width of each latent with resolution 4 or greater by sampling random Gaussian noise for the new dimensions; this gaussian noise can be considered a sample from our standard normal prior. We leave the $1 \times 1$ latent intact to ensure the new regions share global context of the original image. These new latents are then decoded to produce an image.

\begin{figure}
\begin{center}
\includegraphics[scale=0.1]{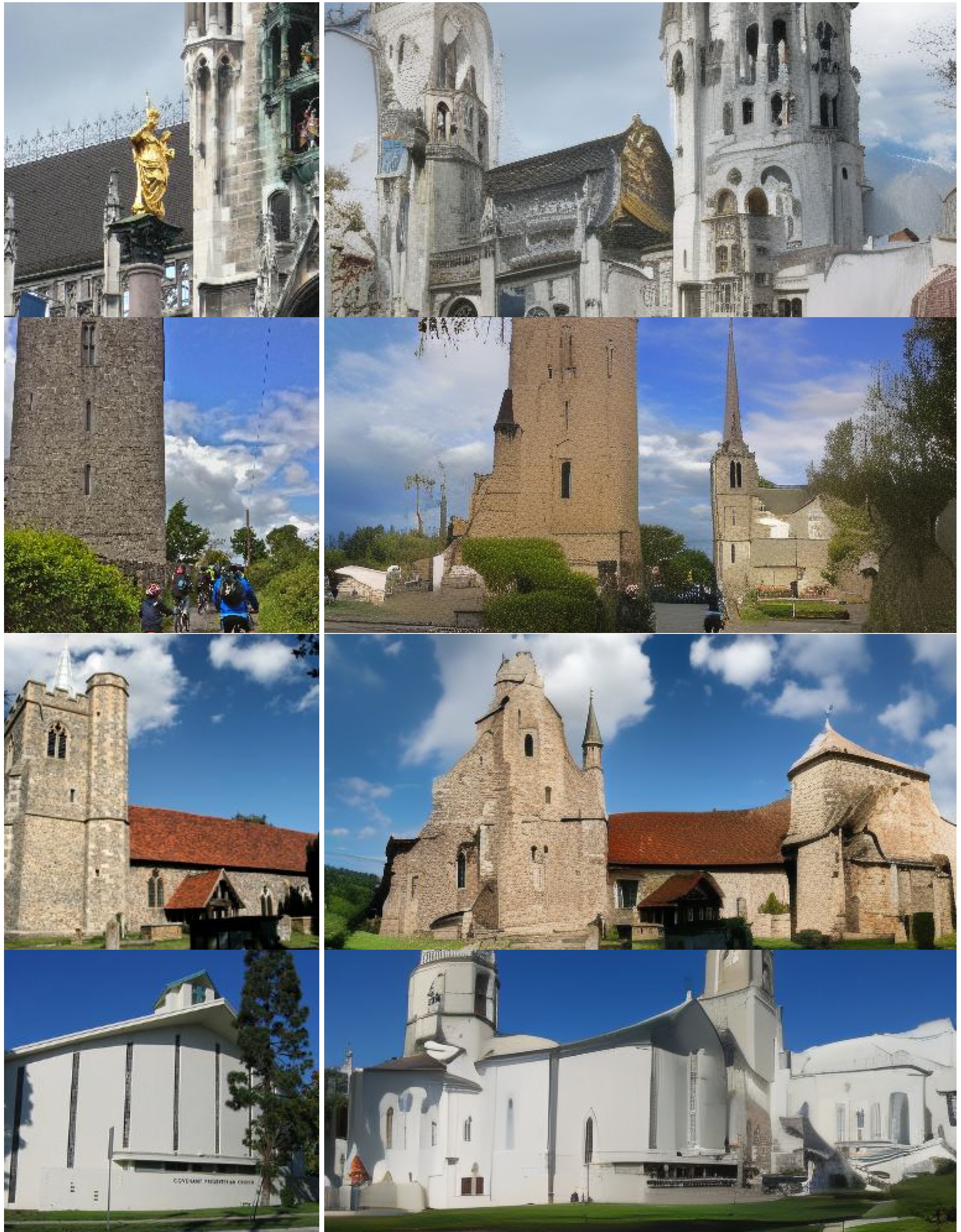}
\caption{The left column shows images from the dataset, and the right shows outpainted samples from our model.}\label{fig:6}
\end{center}
\end{figure}

Figure \ref{fig:6} visualizes our outpainting results. The model can generate convincing outpainted regions that are smoothly integrated into the original image. This indicates impressive generalization abilities considering our model was not trained explicitly for this task, and has not seen any images of this resolution during training. While the samples are not perfect and modify the original image in some places, it nevertheless highlights the usefulness of VAE representations in downstream tasks. Such useful representations, combined with easy access provided by the inference model, extend VAEs' practical applications for endeavors beyond simply generating samples. 

\section{Conclusions}
\label{section:6}

We have demonstrated how hierarchical VAEs trained in latent space can generate highly realistic images on complex multimodal datasets. This performance, combined with their fast sampling, stable training, and flexibile manipulation make them an exciting class of generative models. Furthermore, scaling to large text-to-image datasets might have great benefits for VAEs, the way it has for other likelihood-based models.

One drawback of our method compared to pixel-space VAEs is the inability to compute the data likelihood, which is useful for tasks such as density estimation and anomaly detection. Additionally, our VAEs are still outperformed by best-in-class diffusion models and GANs, particularly in the unguided setting. We believe that further advancements in VAEs might help close this gap.

%\section*{Acknowledgements}

%\textbf{Do not} include acknowledgements in the initial version of
%the paper submitted for blind review.

%If a paper is accepted, the final camera-ready version can (and
%probably should) include acknowledgements. In this case, please
%place such acknowledgements in an unnumbered section at the
%end of the paper. Typically, this will include thanks to reviewers
%who gave useful comments, to colleagues who contributed to the ideas,
%and to funding agencies and corporate sponsors that provided financial
%support.


\bibliography{references}
\bibliographystyle{icml2023}

%%%%%%%%%%%%%%%%%%%%%%%%%%%%%%%%%%%%%%%%%%%%%%%%%%%%%%%%%%%%%%%%%%%%%%%%%%%%%%%
%%%%%%%%%%%%%%%%%%%%%%%%%%%%%%%%%%%%%%%%%%%%%%%%%%%%%%%%%%%%%%%%%%%%%%%%%%%%%%%
% APPENDIX
%%%%%%%%%%%%%%%%%%%%%%%%%%%%%%%%%%%%%%%%%%%%%%%%%%%%%%%%%%%%%%%%%%%%%%%%%%%%%%%
%%%%%%%%%%%%%%%%%%%%%%%%%%%%%%%%%%%%%%%%%%%%%%%%%%%%%%%%%%%%%%%%%%%%%%%%%%%%%%%
\newpage
\appendix
\onecolumn

\section{Introduction to Variational Autoencoders}
\label{appendix:A}

\subsection{Single group VAEs}
\label{appendix:A.1}

For completeness, we provide an introduction to single-group and hierarchical VAEs in this appendix.  Variational autoencoders are generative models of the form $p(\x, \z) = p(\x \vert \z) p(\z)$ where latent variables $\z$ are used to estimate a conditional probability of the data $\x$. While the true posterior $p(\z \vert \x)$ is generally intractable, introducing an approximate posterior $q(\z \vert \x)$ enables us to obtain an evidence lower bound (ELBO) on the data log-likelihood:
\begin{align}
\log p(\x) & = \log \int \pxz p(\z) \dz \\
           & = \log \pars{\int \qsingle \frac{\pxz p(\z)}{\qsingle} \dz} \\
           & = \log \pars{\Expect{\qsingle}{\frac{\pxz p(\z)}{\qsingle}}} \\
           & \geq \Expect{\qsingle}{\log \frac{\pxz p(\z)}{\qsingle}}  \\
           & \geq \Expect{\qsingle}{\log \pxz} - \Expect{\qsingle}{\log \qsingle - \log p(\z)}  \\
 & \geq \Expect{\qsingle}{\log \pxz} - \KLdiv{\qsingle}{p(\z)}
\end{align}
where Equation 4 is obtained using Jensen's inequality. In a VAE, the posterior $\qsingle$ and prior $p(\z)$ are usually diagonal Gaussian distributions, which enables us to learn the posterior using stochastic backpropogation. By reparameterizing $\z = \boldsymbol{\mu}_\theta(\x) + \boldsymbol{\epsilon} \odot \boldsymbol{\sigma}_\theta(\x)$ for $\boldsymbol{\epsilon} \sim \mathcal{N}(\textbf{0}, \bold{I})$, we can differentiate Equation 6 with respect to the mean $\boldsymbol{\mu}_\theta$ and variance $\boldsymbol{\sigma}_\theta$ of the posterior.

\subsection{Hierarchical VAEs}
\label{appendix:A.2}

For complex data distributions such as natural images, a single Gaussian latent layer is often not expressive enough to capture important dependencies. One way of improving expressitivity is to factorize latent distributions over multiple groups $\z \coloneqq \{\z_1, \ldots, \z_N \}$, where the prior is defined as $p(\z) \coloneqq \prod_{i=1}^{N} \prior$ and the posterior as $\qsingle \coloneqq \prod_{i=1}^{N} \posterior$. This conditional dependence creates a hierarchy among latent groups where early latents encode global information that is then used by later groups for modeling finer details. 

A hierarchical VAE's main components include a generator network that assembles latents to reconstruct the data and parameterize the prior, an encoder to extract useful representations from the input, and a sequence of posterior blocks conditioned on the encoded input and generator hidden state. Both the encoder and generator are implemented with residual neural networks; this is especially important for the generator as it allows non-Markovian dependencies where future latents depend on a combination of all previous ones. When considering image data, it is helpful to have the generator's hidden state start at low resolution and progressively increase it to reflect a coarse-to-fine generation process \citep{vdvae}.

\newpage

\begin{figure}
\begin{center}
\includegraphics[scale=0.19]{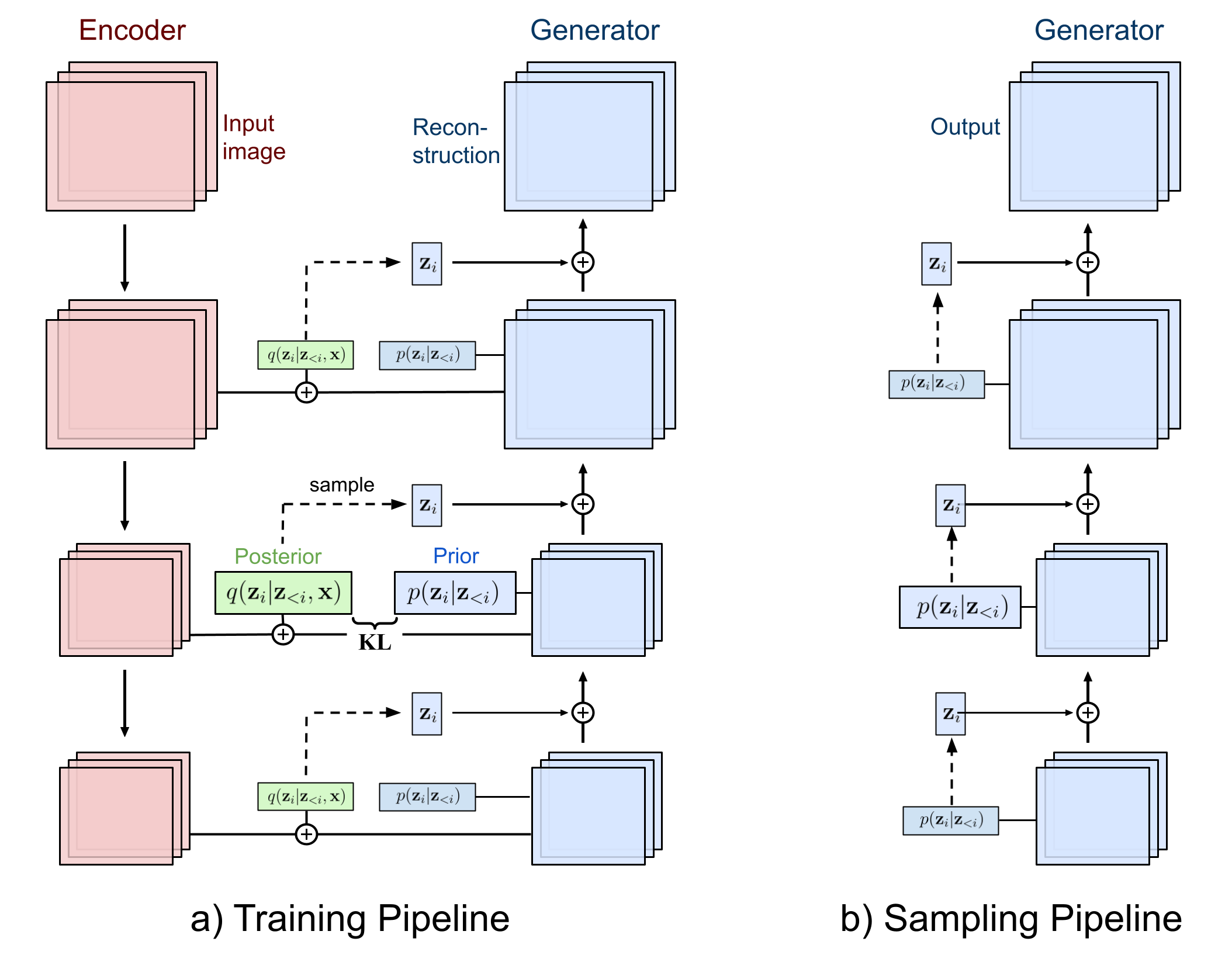}
\caption{A visual depiction of a hierarchal VAE.}\label{fig:7}
\end{center}
\end{figure}

%params: upsample (cascade): 62.8M,  256^2 only: 143.3M
%church small: 119.1M

\section{Implentation Details}
\label{appendix:B}

Our models are based on the deep VAEs from \cite{vdvae}, with several modifications that we detail here. While we did not do official ablations for these changes, we do not believe they have a major positive or negative impact on the model's performance (with the exception of class-conditioning and guidance).

\begin{itemize}
\item{We replace all bottleneck blocks with inverted bottleneck blocks for better memory efficiency. Our inverted bottleneck blocks consist of two $3 \times 3$ convolutions followed by two $1 \times 1$ convolutions in a feed-forward network. }
\item{We introduce multi-headed attention in the encoder and generator for modeling non-local relationships.}
\item{In layers that output prior or posterior distributions, we use spectral regularization \citep{spectralreg}. This is similar to \cite{nvae}, with the difference being they apply it to all layers in the network.}
\item{In our unconditional experiments, we opt to use a (non-hierarchical) standard normal prior instead of a hierarchical one due to its simplicity. We continue to use a hierarchical prior in class-conditional experiments so that the prior is conditioned on labels, and to facilitate guidance in the prior.}
\item{We shift and scale the data by constants so it has means and variances close to zero and one respectively. This normalized data is used as both the input and reconstruction target.}
\item{Similar to \cite{visvae}, our reconstruction loss is a KL divergence between Gaussians instead of a negative log-likelihood loss. The target distribution is centered at the input and has fixed variance $\sigma^2_q$, while the prediction has learned means and variances. This makes it similar to the negative log-likelihood loss, except that it discourages the prior variance from being less than $\sigma^2_q$. The parameter $\sigma^2_q$ controls the strength of the reconstruction term, similar to the $\beta$ parameter in a $\beta$-VAE \citep{betavae}.}
\item{To prevent posterior collapse, we use the KL-scheduling technique from \cite{visvae} which encourages a designated amount of KL to be used in each latent layer.}
\end{itemize}

\begin{table}[h!]
  \begin{center}
    \caption{Training details for ablation and image generation experiments.} 
    \label{tab:table5}
    \begin{tabular}{l c c c c} 
      \\ 
      Dataset & Church Latent-VAE & Church $32^2$ RGB & Church Upsampler & Church Pixel-space \\
      \hline \\
      Parameters & 119.1M & 119.1M & 62.8M & 143.3M \\
      Base channel size & 192 & 192 & 64 & 64\\
      Resolutions & 32,16,8,4,1 & 32,16,8,4,1 & 256,128,64,32 & 256,128,64,32,16,8,4,1\\
      Channel multiplier & 1,1,1,1,1 & 1,1,1,1,1 & 2,2,3,3 & 2,2,3,3,3,3,3,3\\
      FFN expansion ratio & 4,4,4,4,4 & 4,4,4,4,4 & 0,0,4,4 & 0,0,0,4,4,4,4,4\\
      Num latent groups & 8,8,8,8,8 & 8,8,8,8,8 & 3,6,8,8 & 2,5,8,8,8,8,8,8\\
      Num encoder layers & 4,4,4,4,4 & 4,4,4,4,4 & 2,3,4,4 & 2,3,4,4,4,4,4,4 \\
      Num attention layers & 4,4,4,0,0 & 4,4,4,0,0 & 0,0,0,4 & 0,0,0,4,4,4,0,0 \\
      Microbatch size $\times$ \# accums & $128 \times 1$ & $128 \times 1$ & $32 \times 4$ & $32 \times 4$ \\
      Learning rate & $2 \times 10^{-4}$ & $2 \times 10^{-4}$ & $1 \times 10^{-4}$ & $1 \times 10^{-4}$ \\
      $\sigma_q$ & 0.07 & 0.035 & 0.05 & 0.05\\
      SR penalty & 0.1 & 0.1 & 0.25 & 0.25 \\
      EMA decay & 0.9999 & 0.9999 & 0.9998 & 0.9999\\
      Hardware & TPUv2-8 & TPUv2-8 & TPUv2-8 & TPUv3-8\\
      Wall clock time (hr) & 36 & 312 & 345 & 175\\
      
    \end{tabular}
    \begin{tabular}{l c c c c} 
      \\ 
      Dataset & Bedroom f-4 & Bedroom f-8 & Bedroom f-16 & ImageNet-256\\
      \hline \\
      Parameters & 96.2M & 107.2M & 99.8M & 386.4M \\
      Base channel size & 64 & 192 & 192 & 256\\
      Resolutions & 64,32,16,8,4,1 & 32,16,8,4,1 & 16,8,4,1 & 32,16,8,4,1\\
      Channel multiplier & 1,1,1,1,1 & 1,1,1,1,1 & 1,1,1,1 & 1,1,1,1,1\\
      FFN expansion ratio & 0,4,4,4,4 & 4,4,4,4,4 & 4,4,4,4 & 4,6,8,10,12\\
      Num latent groups & 8,12,10,8,6,6 & 8,8,7,6,6 & 10,9,8,6 & 10,12,10,8,8\\
      Num encoder layers & 4,6,5,4,3,3 & 4,4,4,3,3 & 5,5,4,3 & 4,5,4,3,3 \\
      Num attention layers & 0,4,4,4,0,0 & 4,4,4,0,0 & 4,4,0,0 & 4,4,4,0,0 \\
      Microbatch size $\times$ \# accums & $32 \times 4$ & $128 \times 1$ & $128 \times 1$ & $128 \times 8$ \\
      Learning Rate & $2 \times 10^{-4}$ & $2 \times 10^{-4}$ & $2 \times 10^{-4}$ & $3 \times 10^{-4}$ \\
      $\sigma_q$ & 0.1 & 0.07 & 0.07 & 0.07 \\
      SR Penalty & 0.1 & 0.1 & 0.1 & 0.1\\
      EMA Decay & 0.9999 & 0.9999 & 0.9999 & 0.9998 \\
      Hardware & TPUv2-8 & TPUv2-8 & TPUv2-8 & TPUv3-8 \\
      Wall clock time (hr) & 104 & 28 & 17 & 630
      
    \end{tabular}
  \end{center}
\end{table}

Table \ref{tab:table5} shows hyperparameters for our experiments. All models are trained with the Adam \citep{adam} optimizer with $\beta_1 = 0.9, \beta_2 = 0.9$ with a linear learning rate warmup for the first 100 steps. We implement our repository in JAX \citep{jax} and train on either TPUv2-8 or TPUv3-8 hardware. Gradient accumulation is used when a batch cannot fit into memory. When interacting with pretrained autoencoder models, we use LDM PyTorch repository at \href{https://github.com/CompVis/latent-diffusion}{https://github.com/CompVis/latent-diffusion}. All metrics are computed with the evaluation script from \cite{adm} at \href{https://github.com/openai/guided-diffusion/tree/main/evaluations}{https://github.com/openai/guided-diffusion/tree/main/evaluations}.

\newpage

\section{Implementation of Classifier-Free Guidance}
\label{appendix:C}
This appendix provides a code implementation template for performing guided sampling from a hierarchical VAE stochastic layer. While our VAEs are conditioned only simple class labels, this guidance strategy can be easily extended to other conditioning variables such as text.

\begin{lstlisting}
# Hierarchical VAE stochastic layer that supports guidance
class StochasticLayer(nn.Module):
	...
	
	def sample_guided(self, h_c, h_u, label, w_mu, w_sigma):
		uncond_label = torch.ones_like(label) * NUM_CLASSES
		
		cond_output = self.block1(h_c, label) 
		uncond_output = self.block1(h_u, uncond_label) 
		pmean_c, plogvar_c, residual_c = torch.split(cond_output, [self.z_dim, self.z_dim, self.hidden_dim], dim=1)
		pmean_u, plogvar_u, residual_u = torch.split(uncond_output, [self.z_dim, self.z_dim, self.hidden_dim], dim=1)
		
		guided_mean = pmean_c + w_mu * (pmean_c - pmean_u)
		guided_logvar = plogvar_c + w_sigma * (plogvar_c - plogvar_u)
		z = guided_mean + torch.randn(pmean_c.shape) * torch.exp(0.5 * guided_logvar) 
		
		z = self.z_projection(z)		
		h_c = h_c + residual_c + z
		h_u = h_u + residual_u + z
		h_c = h_c + self.block2(h_c, cond_label)
		h_u = h_u + self.block2(h_u, uncond_label)
		return h_c, h_u
    
\end{lstlisting}

\newpage

\section{Nearest Neighbor Visualizations}
\label{appendix:D}

\begin{figure}[hbt!]
\begin{center}
\includegraphics[scale=0.39]{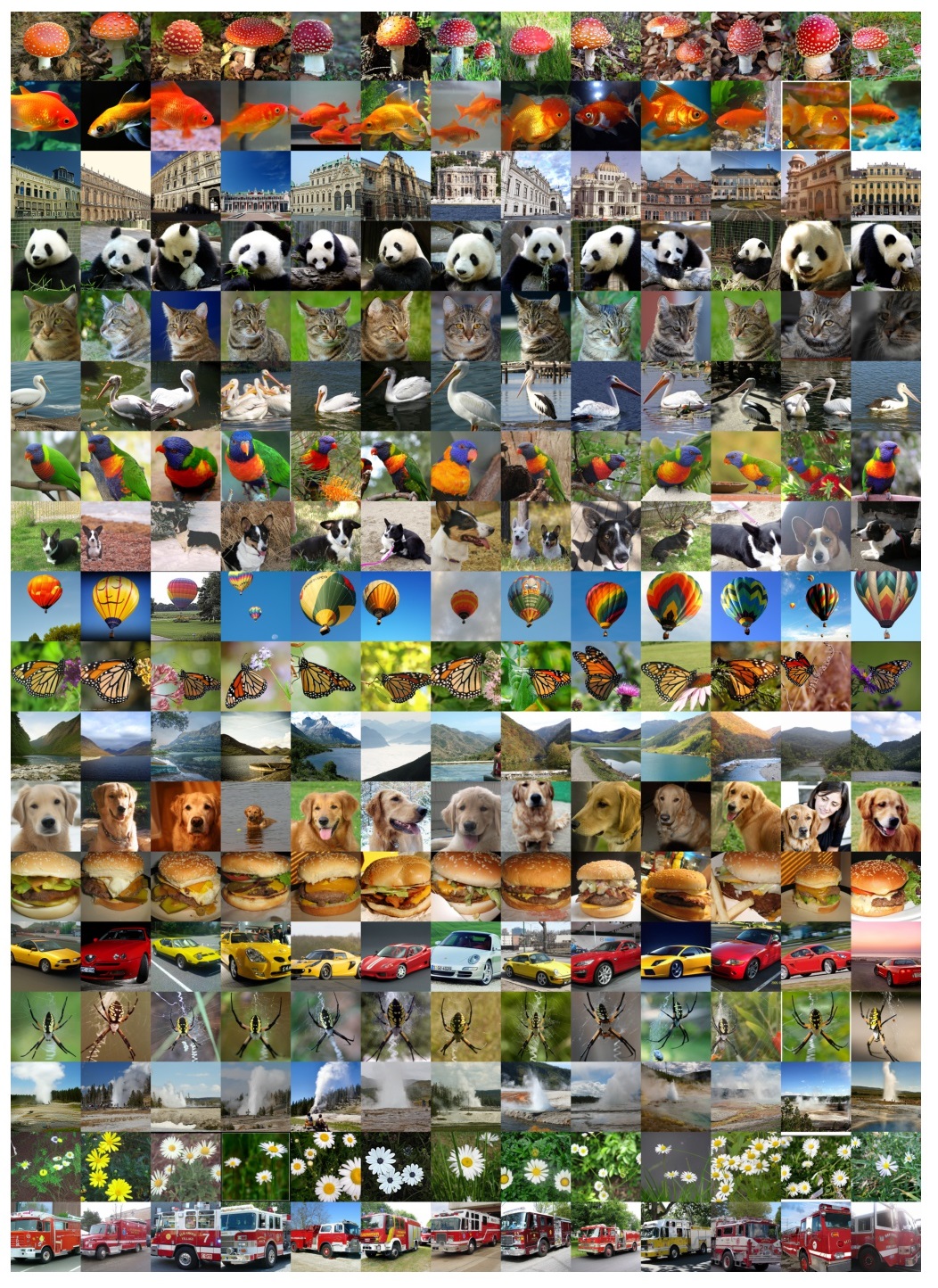}
\caption{Nearest neighbors computed in the feature space of Inception-v3. The leftmost column shows samples from our model, and columns to the right are neighbors from the training set.}\label{fig:nearestfigure}
\end{center}
\end{figure}

\section{ImageNet Samples}
\label{appendix:E}

\begin{figure}[hbt!]
\begin{center}
\includegraphics[scale=0.54]{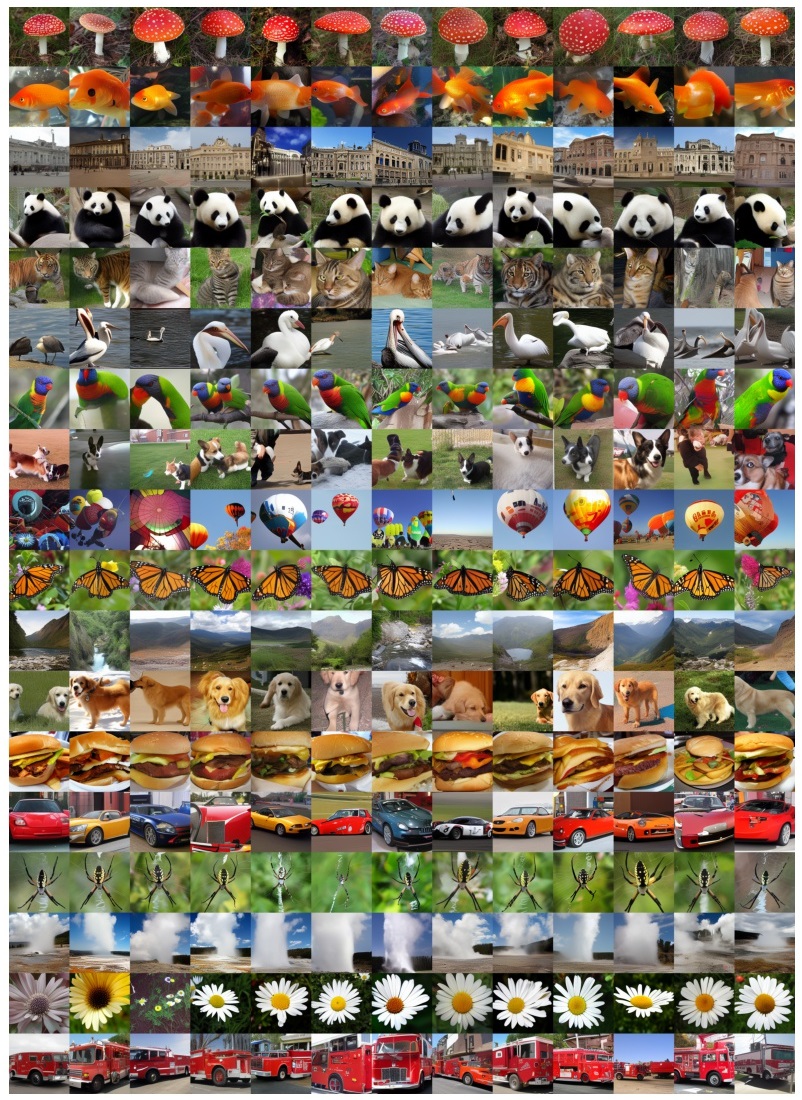}
\caption{Random, guided samples from our ImageNet-256 model (FID=9.34)}
\end{center}
\end{figure}

\begin{figure}[hbt!]
\begin{center}
\includegraphics[scale=0.54]{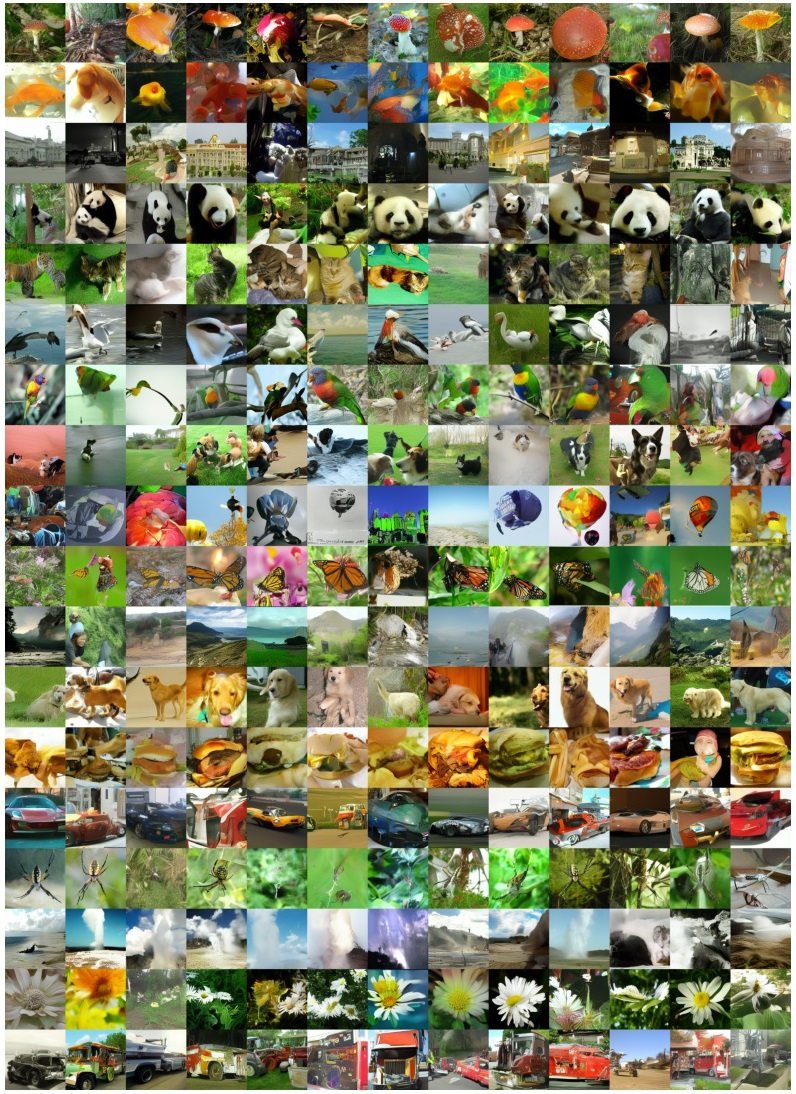}
\caption{Random, un-guided samples from our ImageNet-256 model (FID=32.7)}
\end{center}
\end{figure}

\section{LSUN Samples}
\label{appendix:F}

\begin{figure}[hbt!]
\begin{center}
\includegraphics[scale=0.33]{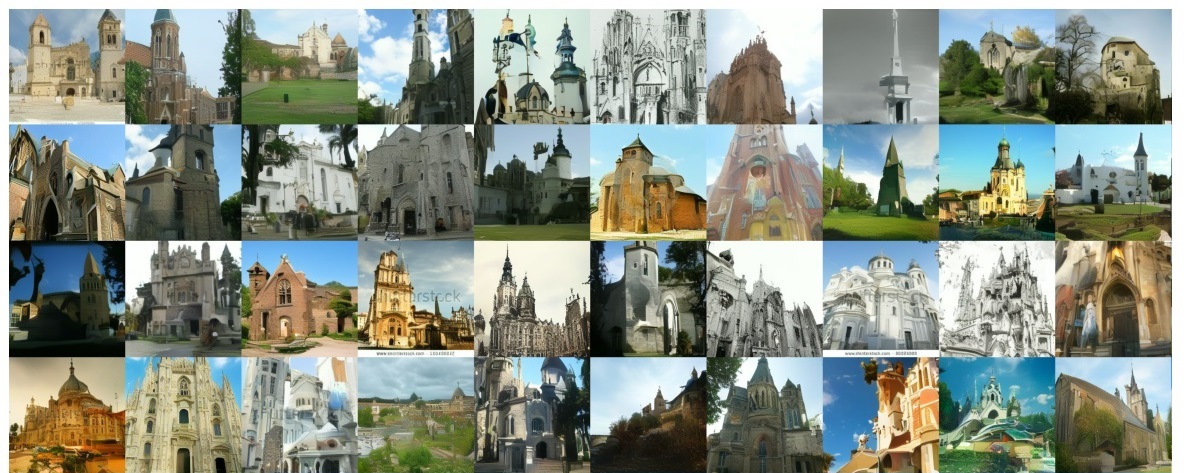}
\vspace{-0.2cm}
\caption{Random samples from our LSUN Church-256 model (FID=7.89)}
\end{center}
\end{figure}

\begin{figure}[hbt!]
\begin{center}
\includegraphics[scale=0.54]{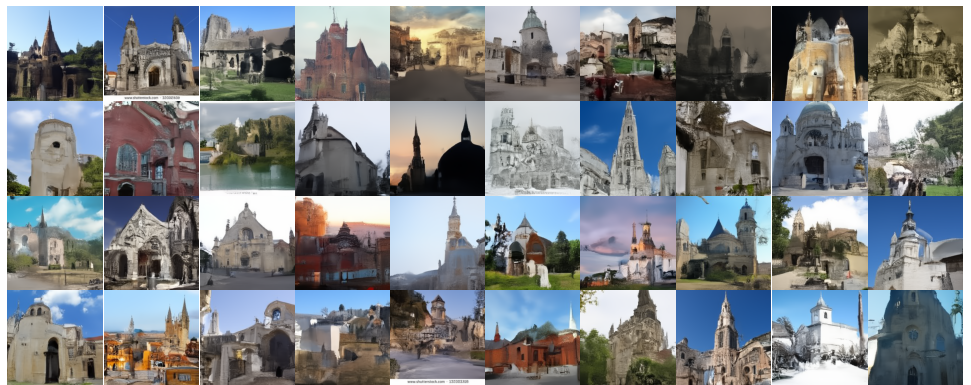}
\vspace{-0.2cm}
\caption{Random samples from a church upsampling stack in Section 4.1's ablation (FID=33.53). \\ Samples exhibit decent large-scale structure but have unrealistic details. It is possible that a\\ more powerful upsampler could improve performance.}
\end{center}
\end{figure}

\begin{figure}[hbt!]
\begin{center}
\includegraphics[scale=0.33]{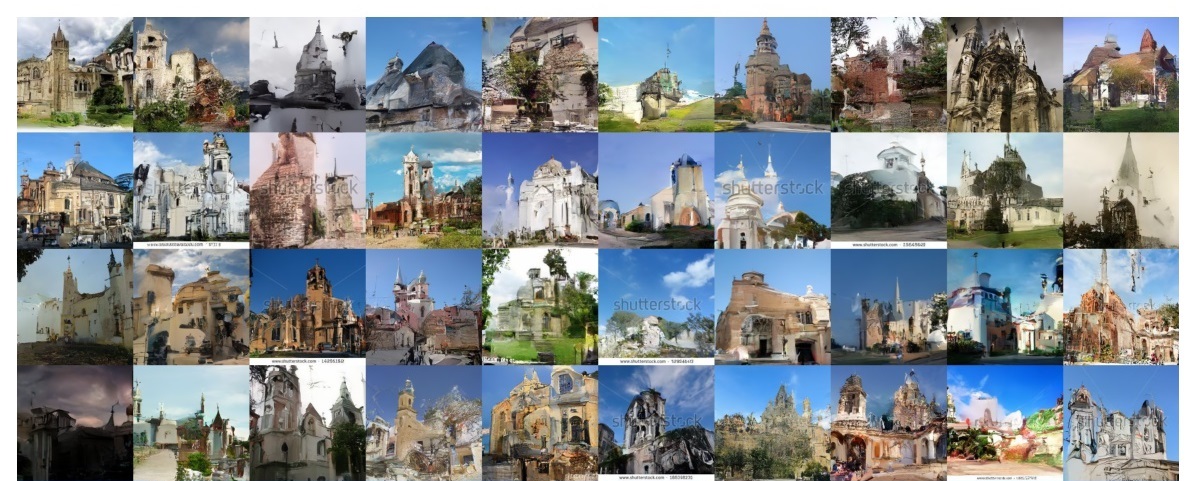}
\vspace{-0.2cm}
\caption{Random samples from a pixel-space model in Section 4.1's ablation (FID=44.36).}
\end{center}
\end{figure}

\begin{figure}[hbt!]
\begin{center}
\includegraphics[scale=0.27]{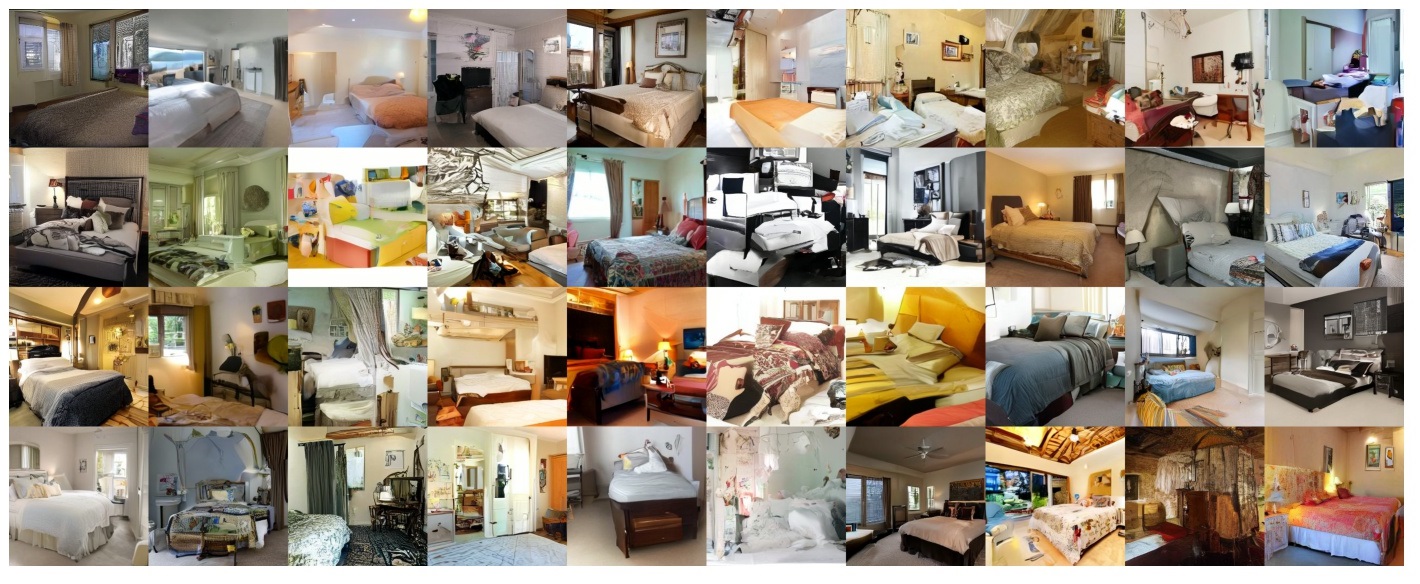}
\vspace{-0.2cm}
\caption{Random samples from our LSUN Bedroom-256 model on a $4\times$ downsampled latent space (FID=9.35).}
\end{center}
\end{figure}

\begin{figure}[hbt!]
\begin{center}
\includegraphics[scale=0.27]{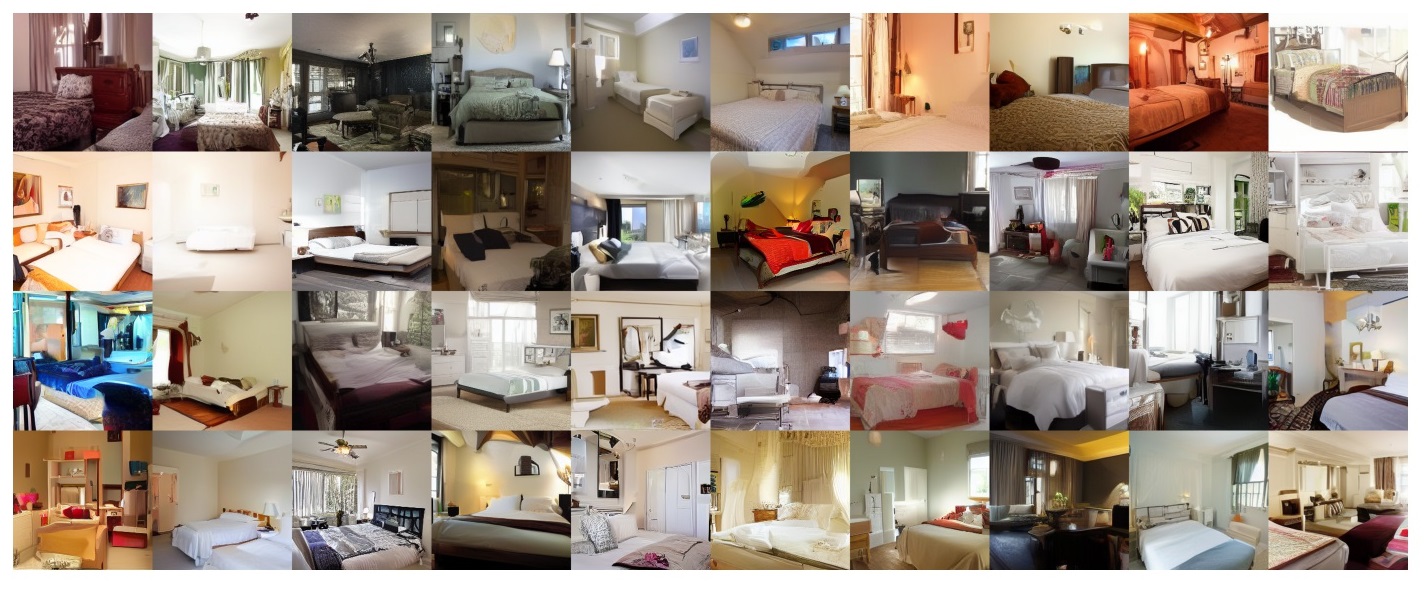}
\vspace{-0.2cm}
\caption{Random samples from our LSUN Bedroom-256 model on a $8\times$ downsampled latent space (FID=11.16).}
\end{center}
\end{figure}

\begin{figure}[hbt!]
\begin{center}
\includegraphics[scale=0.27]{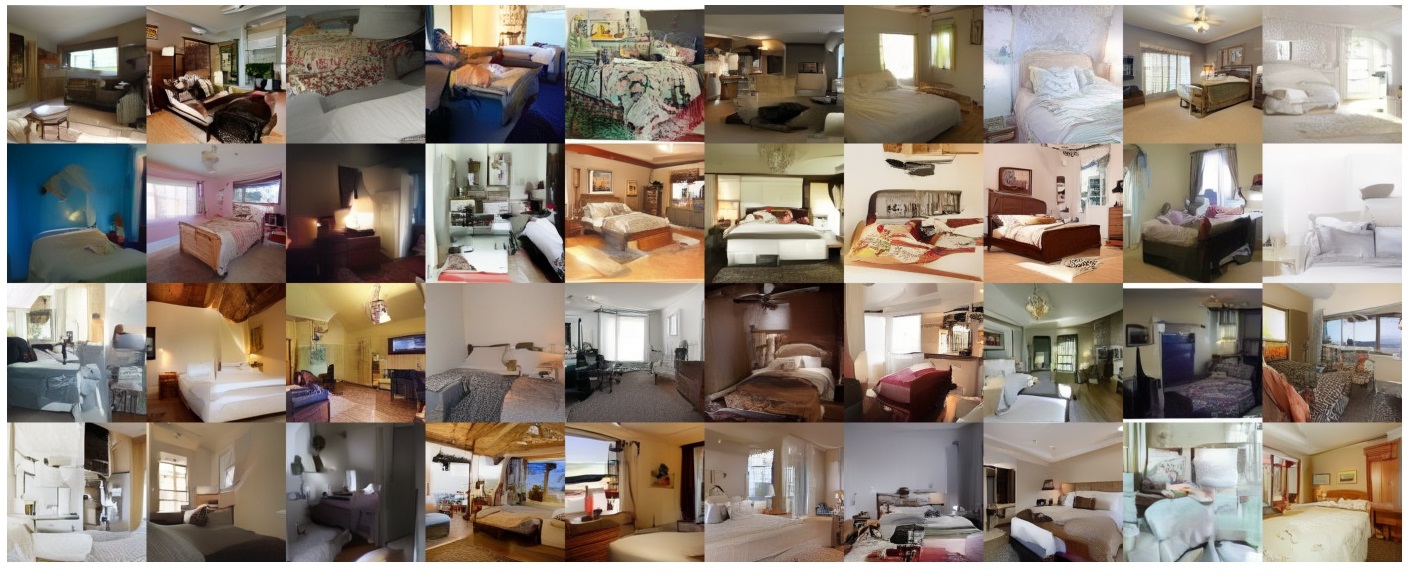}
\vspace{-0.2cm}
\caption{Random samples from our LSUN Bedroom-256 model on a $16\times$ downsampled latent space (FID=17.46).}
\end{center}
\end{figure}

\end{document}